\documentclass{llncs}
\usepackage{makeidx}  
\usepackage{graphicx,subfigure} 
\usepackage{amsmath}
\begin{document}
\frontmatter          
\pagestyle{headings}  
\addtocmark{Hamiltonian Mechanics} 
\mainmatter              
\title{Bayesian Group Nonnegative Matrix Factorization for EEG Analysis}
\titlerunning{Hamiltonian Mechanics}  
%
\author{Bonggun Shin\inst{1} \and Alice Oh\inst{2}}
\authorrunning{Ivar Ekeland et al.} 
%
%

\institute{Korea Advanced Institute of Technology, Daejoen, South Korea,\\
\email{bgshin@kaist.ac.kr} 
\and 
Korea Advanced Institute of Technology, Daejoen, South Korea,\\
\email{alice.oh@kaist.ac.kr} }


\maketitle              

\begin{abstract}

We propose a generative model of a group EEG analysis, 
based on appropriate kernel assumptions on EEG data.
We derive the variational inference update rule 
using various approximation techniques.
The proposed model outperforms 
the current state-of-the-art algorithms
in terms of common pattern extraction.
The validity of the proposed model is tested on
the BCI competition dataset.

\end{abstract}
\section{Introduction}

Electroencephalography (EEG) is a multivariate time-series recording of 
electrical potentials induced by ionic flows among neurons in the brain.
Since EEG has the highest temporal resolution among other non-invasive brain imaging techniques,
it is widely used in the brain computer interface (BCI) research,
especially on the applications where realtime capability is required,
such as  controlling a computer cursor \cite{wolpaw2004control},
mobile robots \cite{millan2004noninvasive}, wheelchair \cite{galan2008brain,shin2010non},
and a humanoid robot \cite{bell2008control}

There have been many approaches to classify mental state based on the 
preprocessed EEG signals.
They include SVM \cite{blankertz2002classifying}, L1 regularized logistic regression \cite{grosse2009beamforming}, 
and nonnegative matrix factorization (NMF) \cite{lee2006nonnegative}.
According to \cite{lee2006nonnegative}, 
NMF based methods do not require any cross-validation
in determining basis vectors which contain useful spectral traits
 in motor imagery EEG signals.

For each mental state, brain images consist of 
subject-dependent patterns
and common patterns shared across multiple subjects.
Most approaches proposed in the literature 
had not considered the latter.
Since those methods can not capture common features occurring across all subjects,
a pilot training phase is always required whenever a new subject comes to the system.
To deal with this limitation, group-NMF \cite{Lee:2009vc} (GNMF) was proposed 
by modifying the cost functions of the standard NMF. 
The advantage of group analysis of EEG is twofold.
First, it finds common patterns that can be used in the testing phase of other subjects without each pilot test,
and second, it finds individual patterns that reflect intra-subject variability.

Most NMF algorithms, including GNMF, are based on
optimization of the cost function under some constraints on the variables.
Although non-generative models give more accurate results in general,
we can not incorporate prior knowledge into them. 
It is well known that EEG data can be well represented with exponential distribution
\cite{zaveri1992time},
but there is no mean to exploit this valuable information in the non-generative models of NMF.
In addition, 
the non-generative models 
is not robust to the small size of data.
while generative models are capable of embedding prior knowledge, 
and competitive performance can be achieved with little data.

With this motivation, 
we devise a generative mode of group EEG analysis, 
based on Bayesian nonnegative matrix factorization.
We derive the variational inference update rule 
using various approximation techniques.
The validity of the proposed model is tested
on the BCI competition III dataset \cite{blankertz2006bci}.

\section{Model Description}


We use preprocessed EEG signals applied power spectral density to have data matrix 
$X \in R^{l \times m\times n}$ 
for each subject $l$.
Each $m$ dimension in $X$ represents frequency bin, and each $n$ dimension is associated with time stamp.
In general, the NMF \cite{seung2001algorithms} finds a decomposition represented as $X = AS$.
However, we assume two kinds of base matrices.
One is common base matrix, $A_C \in R^{m \times k}$.
It reflects activated regions and frequency kinds for a specific task class.
And the other one is individual base, $A_I \in R^{l \times m \times j}$.
The individual patterns vary depending on each subject, 
even though the task is the same.
Hence we model $X$ as
$X_{mnl} = \sum_{k=1}^K (A_C)_{mk} (S_C)_{lkn} 
			+ \sum_{j=1}^J (A_I)_{lmj} (S_I)_{ljn})  $,
where $S_C$ represents class indicator, and 
$S_I$ mixes individual factors appropriately.
It is well known that EEG data can be well represented with exponential distribution
\cite{zaveri1992time}, so we can construct a generative process as follows:
\begin{align}
	X_{mnl} |A_C,(A_I)_l,(S_C)_l,(S_I)_l 
		&\sim \textrm{Exponential}(\sum_{k=1}^K (A_C)_{mk} (S_C)_{lkn} 
			+ \sum_{j=1}^J (A_I)_{lmj} (S_I)_{ljn}) \nonumber \\
	(A_C)_{mk}|a
		&\sim \textrm{Gamma}(a,a) \nonumber \\ 
	(A_I)_{mkl}|b
		&\sim \textrm{Gamma}(b,b)\nonumber \\
	(S_C)_{lkn} &= 1(Y_{ln}=k)  \nonumber\\ 
	(S_I)_{ljn} | Y_{ln},c
		&\sim \textrm{Gamma} (c_{Y_{ln}},c_{Y_{ln}})
\end{align}

\begin{figure}[htbp]
	\begin{center}
		{\includegraphics[height=3.5cm]{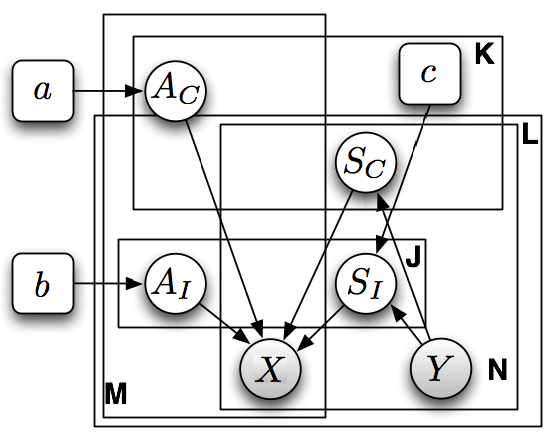}}
		\caption{Graphical Model for Bayesian NMF}
		\label{fig:GMTRAIN}
	\end{center}
\end{figure}
The graphical model for this is shown in Fig. \ref{fig:GMTRAIN}.
We assume gamma distribution for all priors, 
because we can have mathematical advantages of the inference algorithm,
which is shown in Gap-NMF \cite{Hoffman:2010vh}.
We design $A_C$ to have class specific image.
Hence, we assume the number of common bases, $K$, is the same with the number of classes.
The individual bases are designed to be dependent on a subject and a task class.

At the training phase, both $X$ and $Y$ are used as dataset to predict posterior of $A_C,A_I$, 
while $X$ is the only available data in the testing
(Note that we use the posterior of $A_C,A_I$ predicted in the training phase).
For a given estimated posterior of $A_C$,$A_I$, and test data $X^{test}$, 
we predict class label $Y^{test}$ as \\
$ Y_{ln}^{test} = \underset{k}{\operatorname{arg\,max}} \, ||X^{test}_{ln}-\hat{X_{ln}^k}||^2 , \,\,
	\hat{X_{ln}^k} = (A_C)_{k}^{train}$



%

\section{Variational Inference}
There are two kinds of inference techniques in 
 Bayesian graphical models, Markov Chain Monte Carlo (MCMC) and variational inference.
Although MCMC is simple and easy to implement,
it suffers from slow converge speed and no convergence guarantees.
Therefore we derive variational inference algorithm of the proposed model.

We derive variational inference algorithms
 using the similar technique
as introduced Gap-NMF \cite{Hoffman:2010vh} model. 
A typical mean-field variational inference uses 
the same distribution family 
as a variational distribution for each variable,
but Hoffman et al. \cite{Hoffman:2010vh} showed that using Generalized Inverse-Gaussian (GIG) family \cite{jorgensen1982statistical}
as a variational distribution
 gives tighter bound. Therefore we use GIG to approximate $q(A_C)$ and $q(A_I)$.

\subsection{Lower Bound of the Marginal Likelihood}

We can derive the lower bound of the  marginal likelihood of $X$
after we factorize each variable fully.

\begin{align} 
\label{eq:lb}
	\log p(X|a,b,c,Y) 
		&\ge E_q[\log p(X|A_C,S_C,A_I,S_I)] \nonumber \\
			&+ E_q[\log p(A_C|a)] + E_q[\log p(A_I|b)] + E_q[\log p(S_I|c,Y)] \nonumber\\
			&- E_q[\log q(A_C)] - E_q[\log q(A_I)] - E_q[\log q(S_I)]  
\end{align} 

The first term of the bound in (\ref{eq:lb}) can be expanded to (\ref{eq:lik}).

\begin{align}
\label{eq:lik}
	E_q[\log p(X|A_C,S_C,A_I,S_I)] 
		&= \sum_{lmn} \{
			E_q[\frac{-X_{lmn}}{\sum_k  (A_C)_{mk} (S_C)_{lkn}+ \sum_j  (A_I)_{lmj} (S_I)_{ljn}}] \\
			&+E_q[ -\log ({\sum_k  (A_C)_{mk} (S_C)_{lkn}+ \sum_j  (A_I)_{lmj} (S_I)_{ljn}})] \} \nonumber
\end{align}

The first term in (\ref{eq:lik}) can be 
approximated using Jensen's inequality, because $-(\cdot)^{-1}$ is a concave function.

\begin{align} 
	&E_q[\frac{-X_{lmn}}{\sum_k  (A_C)_{mk} (S_C)_{lkn}+ \sum_j  (A_I)_{lmj} (S_I)_{ljn}}]\\
	& \ge - \pi_{lnm1}^2 X_{lmn}(\sum_k \phi_{lmnk}^2 \frac{1}{(S_C)_{lkn}} E_q[\frac{1}{ (A_C)_{mk}}] )
	 	- \pi_{lnm2}^2 X_{lmn} (\sum_j \psi_{lmnj}^2 E_q[ \frac{1}{ (A_I)_{lmj} (S_I)_{ljn}}] )  \nonumber \\
	& \pi_{lnm1}+\pi_{lnm2}=1, \,\, \sum_k \phi_{lmnk}=1, \,\, \sum_j \psi_{lmnj}=1 \nonumber
\end{align}

And for the second term in (\ref{eq:lik}),
we use the same method in \cite{blei2006correlated},
which 
gets lower bound of the convex function, $-\log x$, using 
 a first order Taylor approximation.

\begin{align}
	 E_q[ -\log &({\sum_k  (A_C)_{mk} (S_C)_{lkn}+ \sum_j  (A_I)_{lmj} (S_I)_{ljn}})] \\
	&\ge -\log w_{lmn} +1 -\frac{1}{w_{lmn}} 
		(\sum_k E_q[ (A_C)_{mk} ](S_C)_{lkn} + \sum_j E_q[ (A_I)_{lmj} (S_I)_{ljn}]) \nonumber
\end{align}

\subsection{Optimization}
We present the optimization algorithm 
that maximizes the lower bound in (\ref{eq:lb}),
and it gives the approximated $p(A_C)$ and $p(A_I)$ through $Q(A_C)$ and $Q(A_I)$.

To optimize $\phi$, $\psi$, and $\pi$, we use Lagrange multipliers with sum-to-one constraints.
 
\begin{align}
	\phi_{lmnk}
			& \propto (S_C)_{lkn} E_q[\frac{1}{ (A_C)_{mk}} ]^{-1} ,   \,\,\,\,
	\psi_{lmnj}
			 \propto E_q[ \frac{1}{ (A_I)_{lmj} (S_I)_{ljn}}]^{-1}  \nonumber \\
	\pi_{lmn1}
			&= \frac{\sum_j \psi_{lmnj}^2 E_q[ \frac{1}{ (A_I)_{lmj} (S_I)_{ljn}}]}
			{\sum_k \phi_{mnk}^2 E_q[\frac{1}{ (A_C)_{mk} (S_C)_{lkn}}] +
			\sum_j \psi_{lmnj}^2 E_q[ \frac{1}{ (A_I)_{lmj} (S_I)_{ljn}}]}  
\end{align}

And for the inference of $w$ and other variational parameters,
we use coordinate ascent algorithm to maximize the bound.

\begin{align}
	w_{lmn} &= \sum_k E_q[ (A_C)_{mk} (S_C)_{lkn}] + \sum_j E_q[ (A_I)_{lmj} (S_I)_{ljn}] \nonumber\\
	\gamma_{(A_C)_{mk}}&=a,   \,\,\,\,
	\rho_{(A_C)_{mk}} = a+ \sum_{ln} \frac{1}{w_{lmn}}(S_C)_{lkn} ,   \,\,\,\,
	\tau_{(A_C)_{mk}}
		= \sum_{ln}\{  \pi_{lnm1}^2 X_{lmn}(\sum_k \phi_{lmnk}^2 \frac{1}{(S_C)_{lkn}} ) 	\} \nonumber\\
	\gamma_{(A_I)_{lmj}}&=b,   \,\,\,\,
	\rho_{(A_I)_{lmj}}= b +  \sum_{n}   \frac{1}{w_{lmn}} E_q[(S_I)_{ljn}],   \,\,\,\,
	\tau_{(A_I)_{lmj}}
		= \sum_{n}\{ \pi_{lmn2}^2 X_{lmn}( \psi_{lmnj}^2 E_q[\frac{1}{  (S_I)_{ljn}}] )\} \nonumber\\
	\gamma_{(S_I)_{ljn}}&=c_{Y_{ln}},   \,\,\,\,
	\rho_{(S_I)_{ljn}}= c_{Y_{ln}} +  \sum_{m}   \frac{1}{w_{lmn}} E_q[(A_I)_{lmj}],   \,\,\,\,
	\tau_{(S_I)_{ljn}}
		= \sum_{m}\{ \pi_{lmn2}^2 X_{lmn}( \psi_{lmnj}^2 E_q[\frac{1}{  (A_I)_{lmj}}] )\}
\end{align}

\section{Experiments}

We demonstrate the proposed model on the real EEG dataset.
We compare our model to GNMF, the only group analysis model in the literature.
The performance measure is the classification accuracy.
Throughout the experiments we set all hyper parameters ($a,b,c_1,c_2.c_3$) to 0.1,
the number of common parameter to 3, and the number of individual parameter to 1.

\subsection{IDIAP  Dataset}

The IDIAP  Dataset \cite{blankertz2006bci} is comprised of precomputed features of EEG 
recorded from three subjects.
Each subjects were asked to perform one of the three tasks for some duration of time.
The tasks include imagination of left or right hand movements and 
generation of words beginning with the same random letter.

The preprocessing of the raw EEG is done by 
spatial filtering and power spectral density (PSD).
The raw EEG has 8 centro-parietal channels,
and PSD uses 12 frequency bins at every 62.5 ms,
which constitutes the 96 dimensional feature vector.

\subsection{Common and Individual Factor Extraction}
\begin{figure}[h!] 
	\centering 	\subfigure[Inference of bases of GNMF]
	{\includegraphics[height=3.5cm]{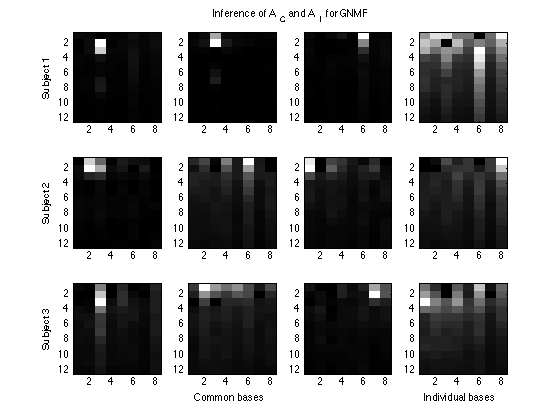} \label{fig:factor1}}
	\centering \subfigure[Inference of bases of the proposed model]  
	{\includegraphics[height=3.5cm]{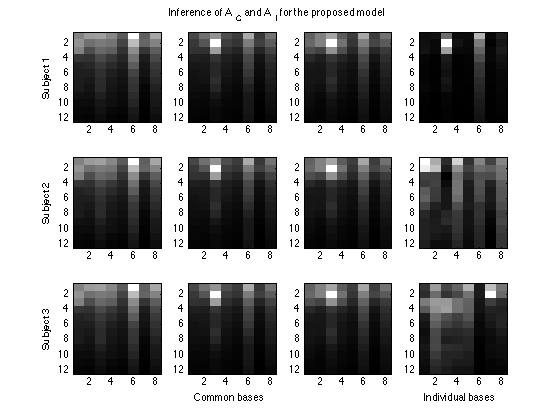} \label{fig:factor2}}
\caption{
According to the result of the BCI competition III,
the best performance was achieved in subject 1,
 which means he or she is less distracted.
Likewise, the subject 3 is more distracted than subject 2.
This fact is well reflected in the proposed model, (b)}
\label{fig:factor}
\end{figure}

In neuroimaging, 
discovering subject independent patterns for a specific task is desirable,
but intra subject variability often thwarts seeking them.
If we can separate the two kinds of patterns,
then the common activation patterns would be more clearly visible.
In Fig. \ref{fig:factor},
we show a side-by-side comparison of the results of GNMF (Fig. \ref{fig:factor1}) 
and our proposed model (Fig. \ref{fig:factor2}) 
in terms of 
the separation of the common and individual bases.
The common bases found by our model are in fact 
common patterns shared by  all three subjects,
whereas the common bases found by GNMF are not quite the same across the subjects.
This shows that our model is better able to separate the common 
patterns from the individual patterns.
Additionally, according to the results of the BCI competition, 
subject 1 showed the best performance, indicating that
he was able to concentrate better on the task than subjects 2 and 3.
Hence, we expect the individual pattern of subject 1 to be clearer 
(more concentrated in a small region) than those of subjects 2 and 3.
The rightmost column of Fig. \ref{fig:factor2}
shows the concentrated pattern around a small region for subject 1 and less
concentrated patterns for subject 2 and subject3.
On the other hand, 
GNMF does not reflect this individual performance difference 
in the individual bases in the rightmost column of in Fig. \ref{fig:factor1}.


\subsection{Sensitivity of the Training Data Size}
\begin{figure}[h!]
	\begin{center}
		{\includegraphics[height=3cm]{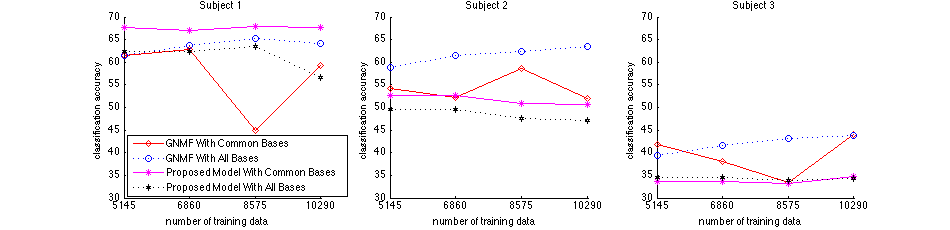}}
		\caption{Performance comparison under various training data size for each subject}
		\label{fig:sensitive}
	\end{center}
\end{figure}

In general, 
the performance of a Bayesian graphical model
is less sensitive to the size of training data
because it can take advantage of the prior.
The proposed model inherits this advantage,
so the performance is robust to the size of the training data.
In practice,
a smaller training dataset is desirable
if it can achieve comparable performance 
because gathering of training data often costs time and money.
Fig. \ref{fig:sensitive} shows
such robustness of our model.
Note that our model performs well with only the common bases (except the subject 3).
This shows that while our model captures the common patterns well, 
it does not capture the individual patterns well. 
This shows the limitation of our model in its current form and 
shows potential for better performance once the model can also capture 
the individual variability.
%
%


\section{Conclusion}

We presented a generative model for analyzing 
group EEG data.
The proposed models finds
common patterns for a specific task class across all subjects 
as well as individual patterns that capture intra-subject variability.
The proposed model seems to capture the common patterns better than 
previously proposed group NMF model,
and it seems less sensitive to the size of the 
training data because it is a generative model.
However,
the limitation of the model is that
it does not model the individual variability well,
and that is left for future research.
We believe that better modeling the individual variability, 
combined with the good performance for common pattern discovery,
will result in an overall improved model.

\bibliographystyle{splncs03}
\bibliography{bnmfMLINI12}

%
%
%
%
%
%
%
\end{document}